\documentclass[linenumbers]{AAS}	
\usepackage{latexsym,amssymb,epsf}
\usepackage{comment}
\usepackage{placeins}
\usepackage{subcaption}
\usepackage[utf8]{inputenc}
\usepackage{textcomp}
\usepackage{xcolor} 
\usepackage{color}
\usepackage{graphicx}
\usepackage{amsmath}
\usepackage{bbm}
\usepackage{amsfonts}
\usepackage[version=4]{mhchem}
\usepackage{url}
\usepackage{siunitx}
\usepackage{longtable,tabularx}
\usepackage[colorinlistoftodos]{todonotes}
\usepackage[font={normal}]{caption}
\newcommand{\kl}[1]{\textcolor{purple}{\textrm{(#1 - KL)}}} 

\usepackage{setspace}
\newcommand{\R}{\mathbb{R}}
\newcommand{\E}{\mathbb{E}}

\newcommand{\cT}{\mathcal{T}}

\newcommand{\cI}{\mathcal{I}}

\newcommand{\cP}{\mathcal{P}}
\newcommand{\A}{\mathcal{A}}

\newcommand{\cV}{\mathcal{V}}

\DeclareMathOperator*{\argmax}{arg\,max}

\newtheorem{helptheorem}{Theorem}{} 

\newtheorem{helplemma}[helptheorem]{Lemma} 
\newtheorem{helpcorollary}[helptheorem]{Corollary} 

\newtheorem{helpexample}[helptheorem]{Example} 

\newtheorem{helpproposition}[helptheorem]{Proposition} 

\newtheorem{helpremark}[helptheorem]{Remark} 

\newtheorem{helpdefinition}[helptheorem]{Definition}

\newtheorem{helpassumption}[helptheorem]{Assumption}

\newtheorem{helpproblem}[helptheorem]{Problem}

\newenvironment{problem} 
{\vskip.1cm\begin{helpproblem}} 
	{\end{helpproblem}\vskip.1cm}

\usepackage{cite}
\usepackage{amsmath,amssymb,amsfonts}
\usepackage{algorithmic}
\usepackage{graphicx}
\usepackage{textcomp}
\usepackage{xcolor}
\usepackage{multicol}

\def\BibTeX{{\rm B\kern-.05em{\sc i\kern-.025em b}\kern-.08em
		T\kern-.1667em\lower.7ex\hbox{E}\kern-.125emX}}

\title{Exposure-Based Multi-Agent Inspection of a Tumbling Target Using Deep Reinforcement Learning
	\footnote{Approved for Public Release; distribution unlimited. Public Affairs release approval AFRL-2023-0295. Work funded under AFRL Contract FA9453-21-C-0602.}}

\author{ 
	Joshua Aurand\thanks{Machine Learning Engineer - Robotics, Verus Research 6100 Uptown Blvd NE, Albuquerque, New Mexico, 87110},
	Steven Cutlip\thanks{Research Engineer - Controls, Verus Research},
	Henry Lei\thanks{Software Controls Engineer, Verus Research},
	Kendra Lang\thanks{Technical Director - Space and Autonomy, Verus Research},
	\\ and Sean Phillips\thanks{Research Mechanical Engineer, Air Force Research Laboratory, Space Vehicles Directorate, Kirtland AFB, New Mexico.}}

\begin{document}
	
	\maketitle

	\begin{abstract} 
		
		As space becomes more congested, on-orbit inspection is an increasingly relevant activity whether to observe a defunct satellite for planning repairs or to de-orbit it. However, the task of on-orbit inspection itself is challenging, typically requiring the careful coordination of multiple observer satellites. This is complicated by a highly nonlinear environment where the target may be unknown or moving unpredictably without time for continuous command and control from the ground. There is a need for autonomous, robust, decentralized solutions to the inspection task. To achieve this, we consider a hierarchical, learned approach for the decentralized planning of multi-agent inspection of a tumbling target. Our solution consists of two components: a viewpoint or high-level planner trained using deep reinforcement learning and a navigation planner handling point-to-point navigation between pre-specified viewpoints. We present a novel problem formulation and methodology that is suitable not only to reinforcement learning-derived robust policies, but extendable to unknown target geometries and higher fidelity information theoretic objectives received directly from sensor inputs. Operating under limited information, our trained multi-agent high-level policies successfully contextualize information within the global hierarchical environment and are correspondingly able to inspect over 90\% of non-convex tumbling targets, even in the absence of additional agent attitude control. 
		
	\end{abstract}
	
	
	\FloatBarrier
	\section{Introduction}
	\label{section:Intro}
	
	The proliferation of man-made objects in Earth's orbit has dramatically improved the capability of space-based services (internet, imaging, science missions, etc.). This comes at the cost of an increasingly congested environment risking the safety and reliability of the same services to interference and cascading space debris. 
	In partial response to the threat posed by space debris and the complexity of on-orbit servicing missions, NASA has identified the task of autonomously inspecting resident space objects as a key enabling technology for next-generation space missions \cite{Starek16}; specifically for autonomous rendezvous, proximity operations, and docking (ARPOD). 
	Inspection is a necessary precursor to on-orbit servicing in an effort to identify possible damage, locate docking points, and plan an approach if the object is uncontrollable/non-cooperative (e.g., tumbling). 
	Thus, the inspection problem typically requires comprehensive imaging of the target surface necessary for relative state estimation, pose estimation, and inertia estimation. 
	This is then used as input for a large variety of ARPOD missions.

	The inspection task itself faces several challenges, including the need to plan rapidly in a nonlinear, high-dimensional, highly constrained, and uncertain environment. Using a team of satellites to perform the inspection has the potential to increase task efficiency with many existing solutions actively exploiting this; see \cite{Lei22,Bernhard20,Nakka21}. 
	To handle the additional complexities that multi-agent modeling and derivations introduce, most existing solutions must make strong simplifying assumptions.
	
	For example, \cite{Bernhard20} and \cite{Nakka21} restrict their solutions to parking satellites on pre-computed elliptical natural motion trajectories around the target, which can increase both the time to complete an inspection and the number of inspecting agents required to get full sensor coverage of the target. In \cite{Phillips2022}, simplified Clohessy-Wiltshire equations and a Lyapunov-based closed-loop control method were used to converge a multi-satellite spacecraft system to a formation around a chief agent. The articles \cite{oestreich21} and \cite{Dor18} focus on relative state and pose estimation from realistic images, and formulate the inspection task as a simultaneous localization and mapping (SLAM) problem. \cite{oestreich21} does so using a factor graph, and implements their solution on a hardware testbed, while \cite{Dor18} uses a novel technique they call ORB-SLAM. Neither imposes major assumptions on the target, but both assume a fixed trajectory for the inspecting agent rather than including optimization over data collection and/or fuel consumption. When trajectory optimization is explicitly considered, the problem of relative state estimation of the target is often simplified by assuming the target is cooperative with only a single inspecting agent, as in \cite{woffinden07}. 
	In \cite{Maestrini22}, trajectory optimization is combined with an unknown, non-cooperative target but again using only a single inspecting agent. Furthermore, this solution requires onboard trajectory sampling and simulation that quickly becomes computationally expensive with a large potential for excessive power draw.

	We seek to  create a general solution framework that combines optimal pose and relative state estimation for an unknown/non-cooperative target while also optimizing inspecting agent trajectories to reduce fuel consumption or time to complete an inspection, in part through the use of multiple inspecting agents. We argue that tools in machine learning are particularly well-suited to handle the spatiotemporal complexity of inspection planning operations through high upfront computational requirements but low computational demands at deployment. Through the utilization of deep reinforcement learning, we aim to strike a balance between tractability and simplifying assumptions; see \cite{dh93,googleFRL}. In this, diverse formulations of inspection criteria and tumbling dynamics can be accommodated through offline simulation. This allows for the training of a policy that remains computationally feasible for online calculation. It has the distinct advantage of not requiring explicit consideration of the underlying nonlinear dynamics during controller design. The resulting trained policy may also robustly capture a host of diverse behaviors without needing to recompute optimal control inputs in real-time. Reinforcement learning then has the potential to overcome many of the challenges faced when executing complex maneuvers in space; see \cite{BL19,OLG21,RLattitude}. 
	In multi-agent reinforcement learning, however, these challenges are more pronounced and less work has been done in this area; for instance, see \cite{Lei22,WANG2011,taxirl,MAMP}.
	
	This paper presents a multi-agent deep reinforcement learning (RL) solution to learn on-orbit satellite inspection within an extendable and flexible framework. 
	We focus specifically on the decentralized navigation of multiple agents to achieve a certain percentage of inspection coverage through image collection in a fuel- or time-optimal manner for a non-cooperative/tumbling target. 
	We do not consider the problem of pose or relative state estimation (although argue that our RL solution can be extended to encompass this as well), but rather assume a priori knowledge used to generate a point cloud model of the target. Using this, we  assume that agents have access to sensor visibility indices based on target pose, camera range, the field of view (FOV), and agent relative positions. Successful inspection is defined through information retrieval, which is a function of agent and target motion, rather than the traversal of an a priori fixed set of ``inspection points'', as in our previous work \cite{Lei22}. 
	Hence, the time and energy required to successfully inspect the target is variable and well-suited to more complex information-theoretic approaches.
	
	We formalize the problem as a decentralized partially observable Markov decision process (DEC-POMDP). 
	Decentralization captures the need for each agent to execute its own navigation strategy without duplicating or conflicting with the work of the other agents. 
	The POMDP is necessary given the lack of perfect information sharing across agents. 
	We operate under the assumption that agents do not share reconstructed point clouds of the target, and therefore each agent only has partial information about the joint visibility ledger. The DEC-POMDP is solved to optimize two, potentially conflicting, objectives. The first objective is to achieve a combined point cloud reconstruction across all agents that guarantee a fixed proportion of surface coverage (e.g., view at least 90\% of all visible points in the cloud). The second is to minimize a performance metric, which could be time to complete inspection or total agent fuel consumption.
	
	Given the complexity of DEC-POMDPs, we decompose the problem in a hierarchical manner. We fix a number of viewpoints centered around the target that are independent of the target's rotational motion. The agents are limited to traveling between these points, and only take images upon arrival at a viewpoint. Hence the problem may be decomposed into a high-level planner that tells the agent which viewpoint to travel to, and a low-level navigation controller that then carries the agent along a viewpoint transfer. Satellite point-to-point navigation is well understood, and model predictive control (MPC) has often been used for satellite applications; see \cite{petersen21, Morgan14, Foust20}. We utilize a basic velocity-based MPC controller for navigation that supports simple expressions for single-burn estimates of the controlled velocity. This allows us to focus on high-level planning requiring the anticipation of target motion and information retrieval from different viewpoints across highly variable mission time horizons. This is nontrivial due to both target dynamic mode and surface geometry. As such, we employ multi-agent deep RL, specifically the recurrent replay distributed deep Q network algorithm (R2D2, \cite{Kapturowski18}) to construct high-level decentralized policies for each agent. 
	
	We test the composed hierarchical solution in a simulated environment and find that the trained agents effectively balance the exploration of their environment with the minimization of their performance objective. Further, the agents learned to coordinate without explicit communication about target visibility - only broadcasting their own position and velocity. As compared to our previous solution \cite{Lei22} that requires visitation of a full set of inspection points tied to the body frame of the target, our current results demonstrate that navigation through the entire viewpoint space is unnecessary and that dissociating the viewpoints from the target body frame enables efficient inspection of a tumbling target. Finally, the use of deep RL allows us to synthesize policies robust to target motion, in which the agents learn to cooperate and anticipate viewpoints that maximize information retrieval without having to explicitly construct and store a model of the target or the other agent's behavior. Our approach further guarantees inspection coverage up to a user-specified percentage of the target surface area while simultaneously optimizing for time or fuel consumption.
	
	The rest of the paper is organized as follows. Section \ref{section:Background} summarizes the concepts necessary to understand our solution. Section \ref{section:Problem} outlines the problem formulation, including the environment setup and how we decompose the viewpoint selection planner from the navigation controller. Our methodology, including how we set up our problem for a reinforcement learning solution, is given in Section \ref{section:Methodology} followed by results in Section \ref{section:Results} and concluding remarks in Section \ref{section:Conclusions}.

	\FloatBarrier
	\section{Background}
	\label{section:Background}
	Here we briefly document key prerequisite knowledge needed for our analysis. This includes background on space dynamics in Section~\ref{subsec:spaceDynamics} and a brief summary of policy generation utilizing reinforcement learning as a tool to approximate solutions of DEC-POMDPs in Section~\ref{subsection:rl}. 
	
	\subsection{Space Dynamics:}\label{subsec:spaceDynamics}
	\subsubsection{Reference Frames:} 
\begin{figure}[th!]
		\centering
		\includegraphics[width=.8\linewidth]{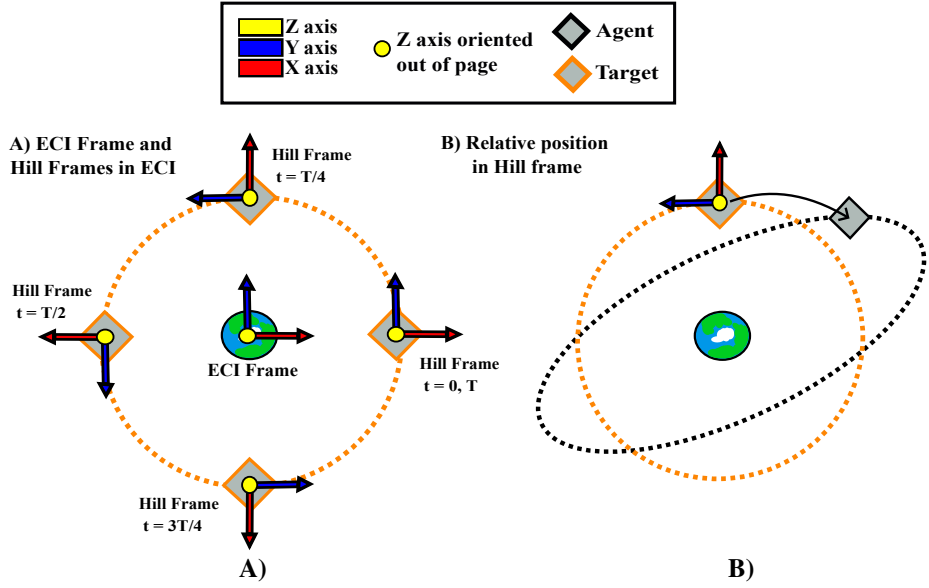}
		\put(-260,-10){\bf A)}
		\put(-75,-10){\bf B)}
		\caption{A) Example unit axes of Hill's frame are shown in the ECI frame at five different time steps with respect to the the orbital period T. B) The position for an agent can be represented as a relative position with respect to Hill's frame. }
		\label{fig:OrbitFrames}
	\end{figure}

The earth-centered inertial frame (ECI) is a global inertial reference frame whose origin is at the center of mass of the earth with a fixed orientation relative to the constellation field. In the ECI frame, the z-axis is close to being aligned with the North Pole but is not incident due to the precession of the earth. The Hill frame (H) is a non-inertial frame fixed on an orbiting body such that the y-axis is tangent to the orbital path, the x-axis points radially away from the earth, and the z-axis completes the orthonormal right-handed coordinate system; seen in Fig. \ref{fig:OrbitFrames}A. Body frames are fixed to each orbiting vehicle and are aligned with their respective principal inertial axes.
	
	\subsubsection{Clohessy-Wiltshire-Hill Dynamics:}
	The orbital path of the inspection target is described using Keplerian mechanics defined through several variables including mean motion, earth radius, J2 perturbation, and orbital radius; see \cite{FMark}. Assuming an orbital eccentricity of zero with no external torques or forces, the relative motion between an agent and the target can be described through Clohessy-Wiltshire-Hill (CWH) dynamics, see \cite{Clohessy60}; an example of target and agent orbit is shown in Fig.~\ref{fig:OrbitFrames}. This permits a linearized form of the CWH dynamics given by:
	\begin{align}
		\dot{\mathbf{x}}(t) =
		\begin{bmatrix}
			0 & 0 & 0 & 1 & 0 & 0\\
			0 & 0 & 0 & 0 & 1 & 0\\
			0 & 0 & 0 & 0 & 0 & 1\\
			3n^2 & 0 & 0 & 0 & 2n & 0\\
			0 & 0 & 0 & -2n & 0 & 0 \\
			0 & 0 & -n^2 & 0 & 0 & 0\\
		\end{bmatrix}
		\mathbf{x}(t)
		+
		\begin{bmatrix}
			0 & 0 & 0 \\
			0 & 0 & 0 \\
			0 & 0 & 0 \\
			\frac{1}{m} & 0 & 0 \\
			0 & \frac{1}{m} & 0 \\
			0 & 0 & \frac{1}{m} \\
		\end{bmatrix}
		\mathbf{u}(t)
		\label{eq:CWH}
	\end{align}
	where $\mathbf{x}(t)$ is a state vector of Euclidean positions and velocities in $\mathbb{R}^6$, $n$ is the mean motion parameter of the target representing its orbital angular frequency, $m$ is the mass of the agent, and $\mathbf{u}(t)=[u_{x},u_{y},u_{z}]^{\top}\in \R^{3}$ are the forces produced by the agent. For simplicity, we assume $u$ is produced by thrusters on the satellite and no additional internal or external forces are present. For a more detailed description of the assumptions required to establish \eqref{eq:CWH}, see \cite{petersen2021} or \cite{wiesel1989}. A family of closed-form solutions, called Natural Motion Trajectories (NMTs), are derived from \eqref{eq:CWH} by setting $u(t) \equiv 0$, computing the Laplace transform, and solving for $x(t)$, $y(t)$, and $z(t)$ explicitly; see the derivation of Eq. 82 in \cite{Irvin2007}).

	
	NMTs can be parametrized by starting position $[x_{0},y_{0},z_{0}]$, ending position $[x_{f},y_{f},z_{f}]$, and a feasible time-of-flight (TOF) $T>0$. Viewpoint transfers are determined by the fixed initial velocity needed by the agent to place them on the parameterized NMT connecting a starting and ending position. We later utilize this to approximate low-fuel impulsive burn commands to navigate between points in Hill's frame. Following Eq. 109 in \cite{Irvin2007}, the required initial velocity is:
	\begin{align}
		\begin{bmatrix}
			\dot{x}_o\\
			\dot{y}_o\\
			\dot{z}_o\\
		\end{bmatrix}
		=
		\begin{bmatrix}
			\frac{-4S+3nTC}{D} &\frac{2-2C}{D} & 0 & \frac{4S-3nT}{D} &\frac{-2+2C}{D} & 0\\
			\frac{-14+6nTS+14C}{D} &\frac{-S}{D} & 0 & \frac{2-2C}{D} & \frac{S}{D} & 0\\
			0 & 0 & \frac{C}{S} & 0 & 0 & \frac{1}{S}\\
		\end{bmatrix} 
		\begin{bmatrix}
			x_0\\
			y_0\\
			z_0\\
			x_f\\
			y_f\\
			z_f\\
		\end{bmatrix} 
		\label{eq:initialvelocity}
	\end{align}
	where $S = \sin(nT)$, $C = \cos(nT)$ and $D = 8-3nTS-8C$. Similarly, the final velocity upon arrival at $[x_{f},y_{f},z_{f}]$ is calculated in eq. 113 of \cite{Irvin2007} as: 
	\begin{align}
		\begin{bmatrix}
			\dot{x}_f\\
			\dot{y}_f\\
			\dot{z}_f\\
		\end{bmatrix}
		=
		\begin{bmatrix}
			\frac{-4S+3nTC}{D} &\frac{-2+2C}{D} & 0 & \frac{4S-3nT}{D} &\frac{2-2C}{D} & 0\\
			\frac{2-2C}{D} &\frac{-S}{D} & 0 & \frac{-14+6nTS+14C}{D} & \frac{S}{D} & 0\\
			0 & 0 & \frac{-1}{S} & 0 & 0 & \frac{C}{S}\\
		\end{bmatrix} 
		\begin{bmatrix}
			x_0\\
			y_0\\
			z_0\\
			x_f\\
			y_f\\
			z_f\\
		\end{bmatrix}. 
		\label{eq:finalvelocity}
	\end{align}
	
	\subsubsection{Torque-Free Rigid Body Dynamics:}\label{subsection:tumbling_dynamics}
	The evolution of attitude dynamics of the inspection target is modeled by Euler's rotational equations of motion for rigid bodies. Supposing the body frame (BF) of the inspection target aligns with the principal axes of inertia and assuming that there are no external applied torques, the equations of motion take the following form: 
	\begin{align}
		I_{xx}\dot{\omega}_x &= (I_{yy} - I_{zz})\omega_y\omega_z\nonumber \\
		I_{yy}\dot{\omega}_y &= (I_{zz} - I_{xx})\omega_x\omega_z\nonumber \\
		I_{zz}\dot{\omega}_z &= (I_{xx}- I_{yy})\omega_x\omega_y
		\label{eq:ERQ_MOI_NoTorque}
	\end{align}
	where $I \in \mathbb{R}^{3\times3} = \text{Diag}(I_{xx}, I_{yy}, I_{zz})$ is the inertia matrix, and $\mathbf{\omega} \in \mathbb{R}^{3} = [\omega_x \ \omega_y \ \omega_z]^T$ is the angular velocity of the inspection target in the ECI frame. As in \cite{FMark}, we may then recover a quaternion representation $q^{BF}_{ECI}$ of attitude dynamics from BF to ECI by solving \eqref{eq:ERQ_MOI_NoTorque}. This is given by:
	\begin{align}
		\dot{q}^{BF}_{ECI} =
		\frac{1}{2}
		q^{BF}_{ECI}
		\otimes 
		\begin{bmatrix}
			0 \\
			\mathbf{\omega} \\
		\end{bmatrix}
		.
		\label{eq:ERQ_att}
	\end{align}
	
\begin{figure}[h!]
		\centering
		\includegraphics[width=.6\linewidth]{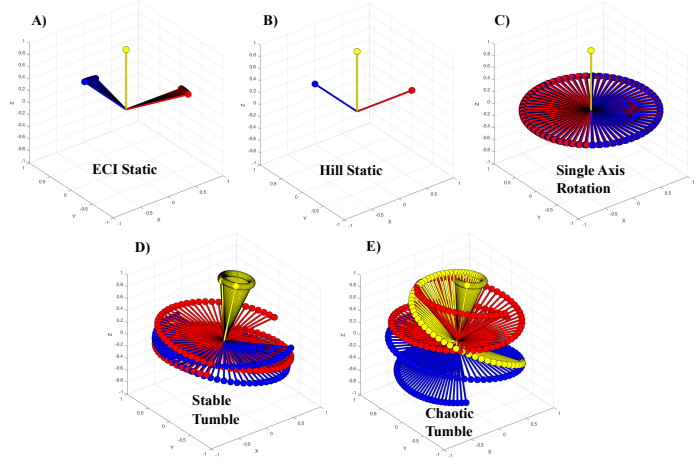}
		\caption{Examples of torque-free rigid body dynamic behaviors as seen in Hill's frame. Each example was generated over a different amount of time to exemplify distinctive qualitative features of each mode.}
		\label{fig:CombinedDynamics}
	\end{figure}

	The resulting dynamics can induce a variety of different behaviors in target attitude evolution, a few examples are shown in Fig. \ref{fig:CombinedDynamics}. These capture the evolution of the target body frame relative to the Hill's frame calculated through a frame transformation applied to \eqref{eq:ERQ_att}. The difference between dynamic modes is a direct result of the intermediate axis theorem \cite{VanDamme,Leine2021} applied to various configurations of initial velocity in \eqref{eq:ERQ_MOI_NoTorque}. Relative to Hill's frame, the \textit{Static ECI} mode represents a slow counterclockwise rotation; the \textit{Static Hill} mode represents no rotation; \textit{Single-Axis} represents a clockwise rotation and the tumbling modes represent multi-axis rotations about stable and unstable body-frame axes respectively. Alternatively, \textit{Static ECI} can be considered a star-pointing mode and the \textit{Hill's static} mode a Nadir-pointing mode.

	
	\subsection{Reinforcement Learning and DEC-POMDPs}\label{subsection:rl}
	
	RL is a subset of machine learning techniques commonly used to solve sequential optimization problems. Heuristically this is done through many, repeated interactions with an environment that provides a reward or penalty for each interaction. Utilizing a pool of collected experience, estimation of the ``best" interactions and their corresponding ``value" can be iteratively improved upon. In well-posed problems, convergence to an analytically derived optimal control is guaranteed. For more complicated environments where analytical options are unavailable, RL methods have proven to be indispensable in planning operations; see for instance \cite{Lin92, Mnih13, Mnih15, fedus20}. The multi-agent autonomous inspection problem frequently falls under this umbrella and we focus our motivation on a specific learning objective within RL, known as Q-learning. To do so, we briefly introduce the basis of our optimization problem in the form of a DEC-POMDP.
	
	A DEC-POMDP may be formally represented as a tuple $(\mathbb{D},S,\mathbb{A},T,\mathbb{O},O,R)$ for which: $\mathbb{D} = \{1,\dots,m\}$ is a set of $m$ agents, $S$ is a set of model states, $\mathbb{A} = \prod_{i\in\mathbb{D}}\mathbb{A}_{i}$ is the set of joint actions, $T$ is a transition probability function, $\mathbb{O} = \prod_{i\in\mathbb{D}}\mathbb{O}_{i}$ is a set of joint observations, $O$ is the observation probability function, and $R:S\times\mathbb{A}\rightarrow\R$ is the joint immediate reward or cost function. Using agent-specific subscripts $i$, if $S$ and $R$ can be factored according to $S=\prod_{i\in\mathbb{D}}S_{i}$ and $R(s,a) = f(R_{1}(s_{1},a_{1}),\dots, R_{m}(s_{m},a_{m}))$ for monotone continuous function $f(\cdot)$, we say the DEC-POMDP is \textit{agent-factored}. Frequently used to cast optimal planning problems under partial information where $R(s,a) = \sum_{i\in\mathbb{D}}R_{i}$, this framework forms the basis of many reinforcement learning problems; see for example in \cite{Mnih15, fedus20, Schaul15}. In these, agents seek to maximize the sum of discounted future rewards accrued through sequential interaction with the environment. The time-preference for reward acquisition is modeled by a fixed discount rate $\gamma\in(0,1)$. An agent's \emph{interaction} is modeled through a policy mapping function $\pi_{i}:O_{i}\mapsto \mathbb{A}_{i}$ (potentially stochastic); a joint policy can then be naturally expressed through $\pi = (\pi_{1},\dots,\pi_{m})$. Formally, the policy mapping functions are composed of two separate components. One that maps $O_{i}\mapsto S_{i}$ through so-called belief state estimation, and the second which maps $S_{i}\mapsto \mathbb{A}_{i}$. Belief state estimation is a crucial complication in many problems where the observation space isn't rich enough to reveal the hidden system state. The RL methods that we implement work directly to estimate both steps simultaneously with two different parameterized neural networks resulting in a single mapping from $O_{i}\mapsto \mathbb{A}_{i}$, as seen above. This is enabled by taking advantage of agent-$i$'s \textit{past experience} contained in the observation space. For the RL algorithm to know what actions are considered ``good" or ``bad", it needs to have a metric to determine the value of a policy in terms of the optimization problem itself; this is enabled through the estimation of a so-called \textit{Q-function}. Collecting past experience\footnote{There are multiple ways of defining sets of observations. These range from a linear sequence of fixed and finite cardinality to collections of observations selected through internal analysis of estimable quantities and functions. The algorithm we use forms this through an \textit{experience replay} buffer.} in a set of observations $\overline{o}_{i} = \{o_{i,k}\}_{k=1}^{K}$ we aim to estimate the \emph{observation-value function} $Q_{i}$ given by:
	\begin{equation}\label{eqn:agentQ_ER}
		Q_{i}^{\pi_{i}}(\overline{o}_{i},a_{i}):=\E^{\pi}\left[\sum_{k=0}^{\infty}\gamma^{k}R_{i}\left(S_{i,t+k+1}^{\pi},a_{i,t+k+1}\right)\big| \overline{O}_{i,t}=\overline{o}_{i},\,a_{i,t}=a_{i}\right],\,\forall i\in\mathbb{D}
	\end{equation}
	where $a_{i,s} =\pi_{i}(\overline{o}_{i,s}^{\pi}),\, \forall s\ge t\text{ and }a_{i,t} = a_{i}$ are the actions determined by each agent's policy $\pi_{i}$; $\overline{O}_{i}$ represent the powerset of all agent-i observations of cardinality less than or equal $k$ occurring before time $t\ge0$. The corresponding \emph{value} of $\pi_{i}$, denoted by $V_{i}^{\pi_{i}}$ is the average of $Q_{i}^{\pi_{i}}$ with respect to the measure induced by $\pi_{i}$, ie. $V_{i}^{\pi_{i}}(\overline{o}_{i}):=\sum\limits_{a\in\mathbb{A}}\pi(a|\overline{o}_{i})Q_{i}^{\pi_{i}}(\overline{o}_{i},a)$. The reward that is received depends on the true value of the hidden state, which is what provides the basis for implicit hidden state estimation. Each agent then seeks to find an optimal policy satisfying:
	\begin{equation}\label{eqn:agentPol}
		\pi_{i}^{*} \in \argmax\limits_{\pi_{i}}V_{i}^{\pi_{i}}(\overline{o}_{i}),\text{ corresponding to some } Q_{i}^{*}.
	\end{equation}
	
	The objective of Q-learning is to learn the solution to \eqref{eqn:agentPol} through maximization of the $Q_{i}$ function in \eqref{eqn:agentQ_ER}, converging to the maximal $Q_{i}^*$; see \cite{Mnih13, Mnih15} for the first neural-network based approach. Due to the break in Markovianity when working in environments with partial information, many RL algorithms leverage a transformed version of \eqref{eqn:agentQ_ER} (directly including a model for belief state transition probability) to ensure applicability of dynamic programming methods for stochastic gradient descent on $Q$-function updates; see \cite{Oliehoek16} for a brief summary. So-called model-free methods don't require this and in general don't require any models for the state transition, observation transition, or belief-state probability distribution. This is particularly advantageous when belief-state probability is difficult to model and is commonly circumvented utilizing a recurrent network to carry forward information through time; see for instance the implementation of IMPALA, PG, PPO, and R2D2 in \cite{RLlib}. This is typically balanced with off-policy methods where $Q_{i}^*$ can be directly estimated using a target policy instead of the action policy; key to establishing many early convergence results in RL. For a high-level review of this topic, see \cite{Sutton18}.

	\FloatBarrier
	\section{Problem Formulation}
	\label{section:Problem}
	The autonomous inspection task considers the inspection of a target object in space via one or multiple inspector satellites equipped with sensors that perform surface area scans within a certain range of the target. Inspection completion is measured by cumulative agent sensor exposure to a user-defined proportion of the target surface, as measured through a visibility calculation on predefined points of interest (POIs) on the surface of the target. The inspection process is decomposed into three distinct components: 1) Viewpoint planning (where to take images), 2) Agent Routing/Path Planning (which agents should move to which viewpoints), and 3) Viewpoint transfers ( point-to-point navigation). This takes place in a global (i.e., shared by all actors) environment consisting of feasible agent viewpoints, target POIs, as well as all necessary parameters to fix agent and target dynamics. Our model assumptions and definitions used to construct this are detailed below in Section~\ref{subsection:Env}. Viewpoint planning and vehicle routing are considered in a single RL-driven agent factorized DEC-POMDP henceforth referred to as ``high-level" planning. We implement a simple analytical MPC controller built around the structure of high-level operations to address (3), henceforth referred to as ``low-level" planning.
	
	\subsection{Environment}\label{subsection:Env}
	The global environment in which the inspection task is carried out consists of the following key components: a viewpoint graph, inspecting agents, the inspection target, agent camera operation or point visibility, and initializing parameter specifications. 
	
	\subsubsection{Viewpoints:} 
\begin{figure}[h!]
		\centering
		\includegraphics[width=.9\linewidth]{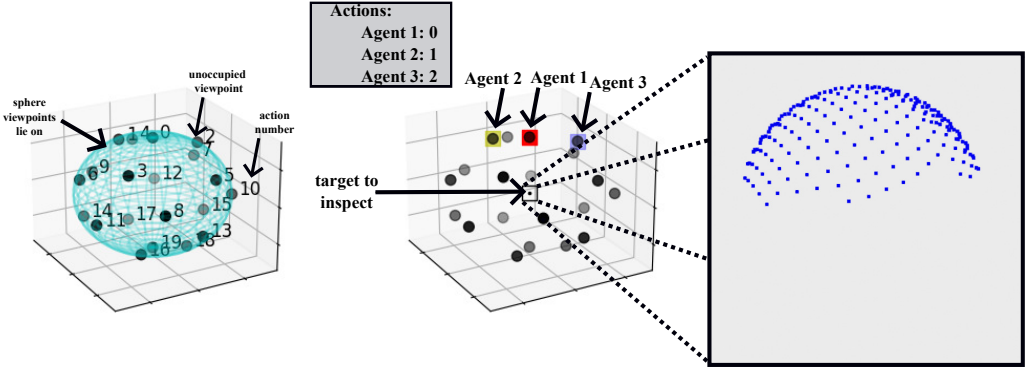}
		\caption{Left: Map of actions to viewpoints. Right: An example of a joint agent action and the resulting collection of visible POIs.}
		\label{fig:ExSnap}
	\end{figure}

We fix a set $\cV$ of $m=20$ viewpoints that stay static within Hill's frame. These represent feasible positions from which agents may take an image of the target. We position them pseudo-uniformly on the surface of a sphere of radius $200m$ via a projected Fibonacci lattice, see \cite{Hardin2016}. This viewpoint generation methodology easily allows us to change the number of viewpoints and provides a simple proxy for reasonable surface coverage of an interior target centered at the origin of Hill's frame. We assume that there are three inspecting agents giving $\mathbb{D}=\{1,2,3\}$ where each high-level planning action represents a viewpoint transfer. We then take the high-level agent-factored action space as the index set of $\cV$, denoted by $\mathbb{A}_{i}=I_{\cV}$. The set $\cV$ is shown in the leftmost plot within Fig. \ref{fig:ExSnap} with labels indicating the action in $I_{\cV}$ that would take an agent to the fixed corresponding viewpoint. A joint action is then a triplet of viewpoint indices corresponding to physical locations in Hill's frame to transfer to.

	\subsubsection{Point Visibility:}
	Upon arrival at a requested viewpoint transfer, each agent takes an ``image" of the target. We assume that each agent maintains a target-pointing attitude so that its camera is pointing towards Hill's frame origin. Each image is captured using a basic FOV filter (set to $15^{\circ}$ ) applied to the set of all visible points given a fixed target attitude and agent position. Point occlusion was handled through reflected spherical projections, as described in \cite{Katz07}; natively implemented by Open3D \cite{open3d}. An example of a joint snapshot by three agents of a unit sphere target is shown in the bottom right portion of Fig.~\ref{fig:ExSnap}. We tuned the diameter of our spherical projection used for hidden point removal to 208874.855 (m).

	\subsubsection{Target Geometry and Dynamics:} 
\begin{figure}[h!]
		\centering
		\includegraphics[width=1\linewidth]{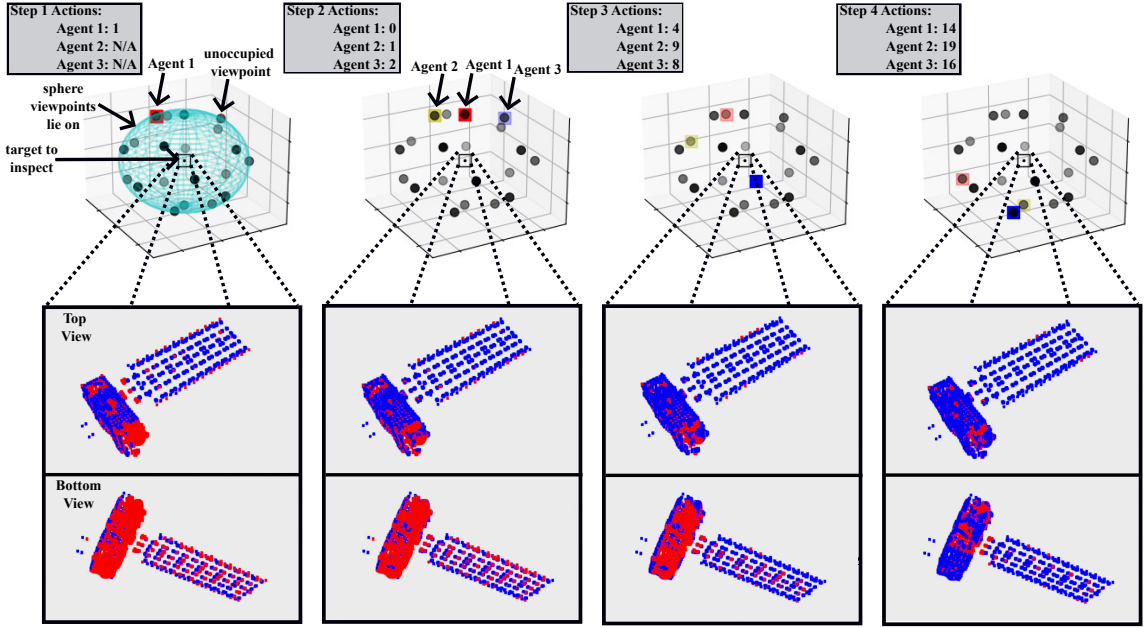}
		\caption{Example inspection of a point cloud in the shape of the Aura Satellite used for training. Step 1 shows an action for a single agent where the discretized sphere of all potential viewpoints is shown. The target to inspect lies at the center of the viewpoint sphere. Steps 2 through 4 show three actions taken by three agents and the corresponding accumulated POIs seen.}
		\label{fig:AuraInsEx}
	\end{figure}

The target geometry used was a point cloud (.ply) representation of the Aura satellite (https://nasa3d.arc.nasa.gov). This contained a full 3-D deconstruction of the satellite including internal components. As these aren't externally visible to our agents, we downsampled a version using Open3D ensuring that 95 percent of the reduced set of points (9514 POIs) were lying on a \textit{visible} target surface. Due to our attitude-pointing assumption, this limits visibility for two dynamic rotation modes (static in H and static in ECI) to a maximum of 93\% of the downsampled point cloud. In particular, the maximum inspection percentage that is achievable across all rotation modes is $<98\%$. This point cloud contained the final set of POIs used for our inspection, which we denote by $\cP$. Inspection progress is tracked utilizing the index set of $\cP$ denoted by $I_{\cP}$. An example inspection of $\cP$ is shown in Fig. \ref{fig:AuraInsEx}, represented by a sequence of joint agent actions. Our inspection threshold used during training was 85\% of $\cP$, corresponding to $\ge 90\%$ of visible POIs. 

	For simplicity, we assume the target is in an orbit with an inclination and eccentricity of zero with no external disturbances. The earth is specified as having a radius $r$ of $6,378.14$km with a gravitational parameter of $3.986004418e^{-5}$ $\frac{km^{3}}{s^{2}}$. The target has an orbital radius $r_{t_{orbit}} = 7357$km. This yields a mean motion of the target equal to $n=0.001027 \frac{1}{s}$ and an orbital period of 6117.99s (1.7hrs). We assume the target has moments of inertia $I_{xx} =$ 100, $I_{yy} =$ 50, and $I_{zz} =$ 70 with an initial attitude $q^{BF}_{ECI} = [1 \ 0 \ 0 \ 0]$. Each dynamic mode tested is determined by the initial angular velocity in the target body frame, denoted by $\omega^{BF}$; these are illustrated in Fig. \ref{fig:CombinedDynamics}.
	
	\subsubsection{Viewpoint Transfers: }
	The fuel cost and transfer time associated with movement between viewpoints (along an NMT) are calculated according to a basic heuristic allowing for an analytical connection between costs for high-level planning and physically realizable trajectory generation during low-level navigation. This segregates high-level from low-level planning and greatly expedites training. It also helps prevent information leakage and reduces the difficulty of causal attribution during training; see Section~\ref{section:Methodology} below for a more detailed discussion. Viewpoint transfers are calculated through single burst thrust targets needed to place an agent on an NMT connecting the viewpoints under a parametrized and predetermined TOF. We assume agents move between viewpoints at roughly the same rate needed to maintain a \emph{closed} NMT or natural motion circumnavigation (NMC). This is determined by the angle between viewpoints in Hill's frame. Parking actions are unique since the natural TOF for an NMC is simply one orbital period. This is overly restrictive since agents may only take an image at the end of a viewpoint transfer. As such, we choose a single burst rule for parking according to 1/2 the time needed to traverse to the nearest neighboring viewpoint. By fixing two viewpoints $v_{1},v_{2}\in\cV$, the TOF is determined by:
	\begin{equation}\label{eqn:LL_proxy}
		\Delta T(v_{1},v_{2}) := 
		\begin{cases}
			\sqrt{r_{0}^{3} / \mu}\arccos\left(\frac{v_{1}\cdot v_{2}}{\|v_{1}\| \|v_{2}\|}\right), & v_{1}\neq v_{2}\\
			\frac{1}{2}\sqrt{r_{0}^{3}/\mu}\min\limits_{\mathbf{x},\mathbf{y}\in\cV}\arccos\left(\frac{\mathbf{x}\cdot \mathbf{y}}{\|\mathbf{x}\| \|\mathbf{y}\|}\right), &  v_{1}=v_{2}
		\end{cases}.
	\end{equation}
	The velocity needed to successfully transfer from $v_{1}$ to $v_{2}$ along an NMT with TOF $\Delta T(v_{1},v_{2})$ is given by $\mathbf{v}_{0}\left(v_{1},v_{2},\Delta T(v_{1},v_{2})\right)$ solving \eqref{eq:initialvelocity}. For a fixed initial velocity at $v_{1}$ given by $\mathbf{v}(v_{1})$ the instantaneous $\Delta V$ required for transfer is then 
	\begin{equation}\label{eqn:delV}
		\Delta V(v_{1},v_{2}) := \|\mathbf{v}_{0}\left(v_{1},v_{2},\Delta T(v_{1},v_{2})\right) - \mathbf{v}(v_{1})\|_{2}.
	\end{equation}
	
	\subsection{Viewpoint Selection and Vehicle Routing}\label{subsection:viewController_prob}

	The viewpoint selection controller or high-level planner is constructed to solve an agent-factored DEC-POMDP determining which viewpoints to visit and in what order. Rewards for the high-level component are determined by point visibility upon arrival at the viewpoint and are penalized on the estimated fuel cost. 
	The joint state space $S:= \A\times\cT\times \cI$ is composed of three distinct parts: agent state $\A$, target state $\cT$, and inspection state $\cI$. The agent state is defined as the combined relative position and velocity of each agent in Hill's frame; $(\mathbf{x}_{i},\mathbf{v}_{i})\in \R^{3}\times\R^{3}$ for $i=1,2,3$ giving $\A:=\R^{18}$. Since the target's translational state defines the Hill's frame origin in ECI, the target's translational dynamics are constant with zero \textit{relative} position and velocity. Target state is then defined solely through attitude and angular velocity with respect to Hill's frame: $(\mathbf{q}_{H}^{BF},\mathbf{\omega}^{H})\in Q\times \R^{3}:=\cT$. Lastly, inspection state is defined through a boolean vector $\mathbf{p} = (p_{1},\dots, p_{N})\in \cI:=\{0,1\}^{N}$ where $N = |I_{\cP}|$ and $p_{i} = 1$ if $i\in I_{\cP}$ has been observed in $\cP$ and is $0$ otherwise. Agent actions in $\mathbb{A}_{i}=\cV$ represent a viewpoint transfer as described in Section~\ref{subsection:Env}. These are determined by a feedback control policy $\pi_{i}(\cdot)$ mapping experience or observation with viewpoint choice. A joint environment step is driven by the simultaneous selection of three actions, one by each agent. More concretely, the $(k+1)^{\text{th}}$ joint environment step is related to agent-level traversal time $t_{i,k+1}$ by:
	\begin{equation}\label{eqn:EnvCalTime}
		t_{k+1} := \max\limits_{i=1,2,3}|t_{i,k+1}| = \max\limits_{i=1,2,3}\big|t_{k}+\Delta T\left(\mathbf{x}_{i}(k),\mathbf{x}_{i}(k+1)\right)\big|.
	\end{equation}
	An important consequence of this is the way in which hidden state information is updated on the target. All agents must make simultaneous planning decisions, even in the face of the sequential rollout. The agent-level state is only updated upon successful viewpoint transfer and incorporates any updates to the inspection state that other agents may have triggered prior to arrival at the commanded viewpoint ($s_{i,t_{k+1}}:=s_{i,k+1}$); the joint update is simply: $s_{t_{k+1}}$. In this model, the state transition function is deterministic. From each state-action pair $(s_{i,k}=s, a_{i,k}=a)\in S_{i}\times \mathbb{A}_{i}$ the witnessing of state $s_{i,k+1} = s'$ is guaranteed with the draw $s'$ determined by the evolution of target dynamics over time $\Delta T(\mathbf{x}^{i},\mathbf{x})$ and forward propagated agent velocity calculated via \eqref{eq:finalvelocity}. 
	
	Agent observations on this space include agent state, target state, a \emph{subset} of inspection states, and traversal time. For instance, fix the $(k+1)^{\text{st}}$ transition state for agent $i$ as $s_{i,k+1}$, as above. Upon arrival at viewpoint $\mathbf{x'}\in \cV$ at time $t_{i,k+1}$, agent $i$ \textit{and only agent $i$} observes $\mathbf{p}_{vis}^{\mathbf{x}'}(t_{i,k+1})$ of the target POIs. This provides an update to the state and forms an observation $o_{i,k+1}$ given by:
	\begin{align*}
		s_{i,k+1} &= [\mathbf{x}_{\{1,2,3\}},\mathbf{v}_{\{1,2,3\}},\mathbf{q}_{H}^{BF}(t_{i,k+1}),\mathbf{\omega}^{H}(t_{i,k+1}),\mathbf{p}_{vis}^{\mathbf{x}'}(t_{i,k+1})\lor \mathbf{p}(t_{j,k+1})],\text{ for }t_{j,k+1}\in[t_{k},t_{i,k+1}]\\
		o_{i,k+1} &= [\mathbf{x}_{\{1,2,3\}},\mathbf{v}_{\{1,2,3\}},\mathbf{q}_{H}^{BF}(t_{i,k+1}),\mathbf{\omega}^{H}(t_{i,k+1}),\mathbf{p}_{vis}^{\mathbf{x}'}(t_{i,k+1}),t_{i,k+1}].
	\end{align*}
	In the above, note that the other agents' positions and velocities are passed in by the last state update $s_{j,k}$ occurring immediately before $t_{i,k+1}$. This prevents agent foresight into future actions and also helps maintain appropriate DEC-POMDP structure. Note that the observation probability function $O$ is determined in much the same way that $T$ for the state space was; deterministic and dependent on dynamics evolution. The state encodes all information that has been previously retrieved by all agents within the inspection, whereas the observation simply encapsulates a single agent's ``picture" of the target. As each agent interacts with the global multi-agent environment, translational and rotational movement are fully communicated but the agent does not witness what others have actually observed of the target.
	
	The immediate reward of a joint state action pair $(s_{k},a_{k})$ is determined through the sum of agent rewards containing a scaled linear combination between exposure based \textit{relative} information gain and expected fuel cost of view traversal. This is defined by:
	\begin{align}\label{eqn:reward}
		r_{i}(s_{i,k}, a_{i,k}) &= 
		\begin{cases}
			\alpha\frac{\|\mathbf{p}(t_{i,k+1})\land\neg \mathbf{p}(t_{i,k})\|_{1}}{|\cP|-\|\mathbf{p}(t_{i,k})\|_{1}} + \beta \Delta V(\mathbf{x}_{k}^{i},\mathbf{x}_{k+1}^{i}) + r_{0}, & \text{ if }|\cP|-\|\mathbf{p}(t)\|_{1}>0\\
			r_{0}, & \text{ if }|\cP|-\|\mathbf{p}(t)\|_{1}=0
		\end{cases}\\
		r(s_{k},a_{k}) &= \sum_{i=1}^{3}r_{i}(s_{i,k}, a_{i,k})\nonumber
	\end{align}
	where $\alpha,\beta,r_{0}$ are fixed real-valued constants and $\Delta V$ satisfies \eqref{eqn:delV}. In our problem, rewards are only accessible to agents while there exists a sufficiently large number of POIs to inspect. This is encoded into the reward function through a random horizon $\tau$ defined by the first hitting time to a user-defined threshold for inspection coverage ratio, call it $M\in(0,1]$. For the inspection state $\mathbf{p}(t_{k})$ at step $k$ and calendar time $t_{k}$, this is defined by
	\begin{equation}\label{eqn:timeHorizon}
		\tau:=\inf\left\{t\ge 0\,,\,\frac{\|\mathbf{p}(t)\|_{1}}{|\cP|}\ge M\right\}.
	\end{equation}
	Note that the law of this random variable is determined by the joint viewpoint selection policy $\pi = (\pi_{1},\pi_{2},\pi_{3})$; therefore becoming a part of the optimization problem. The updated reward functions become:
	\begin{equation}
		R_{i}(s_{i,k}, a_{i,k}) = \mathbbm{1}_{\{t_{k}\le\tau\}}r_{i}(s_{i,k}, a_{i,k}),\text{ and }	R(s_{k}, a_{k}) = \mathbbm{1}_{\{t_{k}\le\tau\}}r(s_{k}, a_{k})
	\end{equation}
	respectively. Putting these together, the viewpoint selection planner aims to solve:
	\begin{problem}[High-Level Planning]\label{prob:VSP}
		For the tuple $(\mathbb{D},S,\mathbb{A},T,\mathbb{O},O,R)$ defining an agent-factored transition and observation independent DEC-POMDP, the viewpoint selection planner aims to find a joint policy $\pi = (\pi_{1},\pi_{2},\pi_{3})$ satisfying \eqref{eqn:agentPol} for each $i=1,2,3$.
	\end{problem}
	The key assumptions we work under as they relate to Problem~\ref{prob:VSP} are:
	\begin{enumerate}
		\item Agents follow an NMT with TOF given by \eqref{eqn:LL_proxy} and approximate fuel cost \eqref{eqn:delV}.
		\item During training, all agents must make simultaneous decisions. Action synchronization is used to ensure a spatial correlation structure isn't formed between agents when training. This is \emph{not} enforced during rollout of the trained policies.
		\item There is no hard constraint preventing the possibility of collision or limiting fuel burn capacity. This was used to help minimize environment complexity but can be readily relaxed. 
		\item All agents are assumed to be target oriented throughout training and camera shutter is linked with viewpoint arrival.
	\end{enumerate}
	
	\subsection{Navigation Control and Hierarchical Policy Rollout}\label{subsection:mpc_prob}
	
	While the high-level planner assigns agents to viewpoints, for implementation, the low-level controller is needed to actually move the agents to those viewpoints. At each simulated time step, the low-level controller is passed an observation tuple consisting of agent position, velocity, goal position and expected traversal time. It solves for the thrust input $u(t)$ in \eqref{eq:CWH} required to achieve a target velocity needed to track the desired NMT. Notationally, we distinguish between two different time scales (high vs low level) using the following notation: $\mathbf{x}_{t}$ denotes agent position in Hill's frame at time $t$, for a feasible time interval $(s-t)>0$ we denote $\mathbf{v}_{0}(\mathbf{x}_{t},\mathbf{x}_{s})$ as the \emph{initial velocity} required to traverse between $\mathbf{x}_{t}$ and $\mathbf{x}_{s}$ over $(t-s)$ TOF whereas $\mathbf{v}_{f}(\mathbf{x}_{t},\mathbf{x}_{s})$ represents the \textit{final velocity}; see eq. (\ref{eq:initialvelocity}) and eq. (\ref{eq:finalvelocity}) respectively. The optimization problem solved between high-level environment steps is formulated as:
	\begin{problem}[Low-Level Planning] \label{prob: OptimizatonLLC}
		Fix a starting time $t_{0}\ge0$, starting position $\mathbf{x}_{0}\in\cV$, target traversal time $t_{f}>t_{0}$, and target viewpoint position $\mathbf{x}_{f}\in\cV$. The initial agent positional state is given by $\mathbf{x}_{t_{0}}=\mathbf{x}_{0}$ and the desired agent control trajectory must satisfy $\mathbf{x}_{t_{f}}=\mathbf{x}_{f}$. For a fixed $n$-step discretization of the TOF $\Delta T:=t_{f}-t_{0}$ with uniform mesh width $\Delta t := \frac{\Delta T}{n}$, the low-level planner solves:
		\begin{align*}
			&\min\limits_{\Tilde{\mathbf{x}},\Tilde{\mathbf{u}}}\sum_{j=0}^{n-1}\|\Tilde{\mathbf{u}}_{t_{j}}\|_{2}\Delta t,\,\text{ such that: } \\
			&\|\mathbf{v}_{0}(\Tilde{\mathbf{x}}_{t_{j+1}},\mathbf{x}_{f})-\mathbf{v}_{f}(\Tilde{\mathbf{x}}_{t_{j}},\Tilde{\mathbf{x}}_{t_{j+1}})\|_{2}=0,\quad \forall j\in\{0,\dots,n-1\}
		\end{align*}
		where $\Tilde{\mathbf{x}}_{t_{j}}$ is the agent's position at time $t_{j}$ under historical thrust control policy $\Tilde{u}_{t_{0}},\dots,\Tilde{u}_{t_{j}}$ evolving according to the discretized form of \eqref{eq:CWH}.
	\end{problem}
	
	Note that since the equality between steps in the constraint is binding, the optimization procedure is therefore implicitly minimizing the slack variable $\Tilde{\mathbf{x}}$ through expected positional drift. This is what ensures consistency with high-level rewards requiring agent position and TOF for information gain and velocity for fuel cost. For comparable approaches on \textit{trajectory} (vs. velocity) constrained optimization, see \cite{Morgan14, Foust20}. 
	
	A solution to Problem~\ref{prob: OptimizatonLLC} represents a specific trajectory for a low level policy performing navigation control. Acting on the space of all feasible viewpoints, this yields a mapping $\pi^{ll}:\R^{9}\times\R_{+}\mapsto \R^{3}$ where the domain represents a low-level observation space consisting of: agent position, agent velocity, agent viewpoint command, and remaining allowable TOF; see \eqref{eqn:obs} below. This is merged with trained high-level policies and used in tandem to perform the multi-agent inspection task. An important consequence of this is the change in environment time scale. Although high-level policies are trained making synchronous decisions according to the environment stepping in \eqref{eqn:EnvCalTime}, when rolling out in the hierarchical environment, regular calendar time is used and the artificial time synchronicity is relaxed. Here, there are two policies for each agent: $(\pi_{i}^{hl}(\overline{o}_{hl}),\pi_{i}^{ll}(o_{ll}))$. At every point in time the agent receives an observation $o(t) = [o_{hl}(t),o_{ll}(t)]$ from the hierarchical environment of the form:
	
	\footnotesize
	\begin{align}\label{eqn:obs}
		o_{hl}(t)
		&= [\mathbf{x}_{\{1,2,3\}}(t^{HL}),\mathbf{v}_{\{1,2,3\}}(t^{HL}),\mathbf{q}_{H}^{BF}(t),\mathbf{\omega}^{H}(t),\mathbf{p}_{vis}^{\mathbf{x}'}(t^{HL}),t^{HL}],\,\text{ if }\mathbf{x}(t_{+}^{HL})\in\cV_{\epsilon}\text{ and } o(t_{-1})=o_{ll}(t_{-1})\nonumber\\
		o_{ll}(t) &= [\mathbf{x}(t),\mathbf{v}(t),\mathbf{x}(t_{+}^{HL}),t_{+}^{HL}-t],\,\text{ if }\mathbf{x}(t_{+}^{HL})\not\in\cV_{\epsilon} \text{ or } |t_{+}^{HL}-t|> \overline{t}.
	\end{align}
	\normalsize
	
	In the above, $t^{HL}$ denotes the calendar time of the agent's current high-level action. $\mathbf{x}(t_{+}^{HL})\in\cV$ denotes a prespecified \emph{goal} viewpoint determined by the high-level agent whereas $t_{+}^{HL}$ denotes the planned time of arrival for the low-level navigator. The set $\cV_{\epsilon}$ denotes a collection of $\epsilon$ neighborhoods around each element in $\cV$; $\overline{o}$ denotes concatenated observations over historical actions stored within the replay buffer. The parameters $\epsilon\in\R_{+}$ and $\overline{t} \in\R_{+}$ represent success criteria imposed on the low level navigator. Each low-level observation provides the information needed to the MPC controller to calculate the subsequent agent control input $u(t)$. Once $\|\mathbf{x}-\mathbf{x}(t_{+}^{HL})\| \le \epsilon \text{ and } |t_{+}^{HL}-t|\le \overline{t}$ are simultaneously satisfied, the next time step will return a high level observation for the viewpoint selection planner which maps to a new goal viewpoint through $\pi_{i}^{hl}(\overline{o}_{hl})$. This continues independently until the prespecified inspection threshold is reached or the environment times out.
	

	\FloatBarrier
	\section{Methodology}\label{section:Methodology}
	
	\subsection{Recurrent Replay Distributed DQN and Environment Crafting}
	
	We solved the high-level planning problem \ref{prob:VSP} utilizing an algorithm known as Recurrent Replay Distributed DQN (R2D2) developed by \cite{Kapturowski18}. It is classified as a model-free, off policy, value-based reinforcement learning algorithm, that is suitable for partially observable environments like the multi-agent inspection problem. The key considerations for this selection were the potential size of hidden state passed through an environment in the form of images or point clouds in addition to the partial observability and proven success in information limited environments.  We chose to limit observations of inspection progress to \emph{just} include only current ``images" of the target and not the collectively observed portion of POIs. In cases where progress is shared but images aren't due to bandwidth or size constraints, if the database is lost or corrupted, the agents will likely lose direction unless dropouts are simulated explicitly during training. Doing so requires an intentional and direct modeling effort that may not be reflective of real-world operation. In this sense, we aimed to provide observations that were as close to those received in ``single-agent" environments while still providing enough information to ensure the multi-agent collaboration remains beneficial to the inspection mission. 
	
	
	Algorithmically, R2D2 follows a similar structure for training as deep Q networks (DQN, \cite{Mnih15}) with modifications to handle recurrent neural networks. The key algorithmic components include: a prioritized distributed experience replay buffer, an n-step double Q learning framework, and an application-specific recurrent deep neural network architecture that is trained through a combination of backpropagation-through-time and stochastic gradient descent. For more details, readers are encouraged to consult \cite{Kapturowski18,Mnih13,Mnih15, Van16,Schaul15, Yu19}. Our training environments were constructed as gym environments to conform with Ray/RLlib's multi-agent training framework, which includes implementations of many of the most popular reinforcement learning algorithms including R2D2; see \cite{RLlib}. Ray's native parameter tuner was then used to find the learning rate, priority exponent, and importance sampling exponent. Each parameter was searched on an interval of $[.9p_{0},1.1p_{0}]$ where $p_{0}$ is the default value used in the standard R2D2 config. Batch size and replay buffer capacity were fixed experimentally to reduce run-time during training; the buffer capacity and max sequence length were determined heuristically as upper bounds for the most action intensive inspection mode, the stable tumble. For each rotational dynamic mode, we trained policies with a common architecture. Observations are pre-processed in accordance to RLlib's default module and fed into a feed forward fully connected network with two hidden layers of size 64 with tanh activation functions. The network is wrapped in an LSTM layer (see \cite{Hochreiter97, Yu19}) with a hidden state of size 64. For a replay sequence of observations and hidden states: $\{o_{i,t+k}\}_{k=1}^{m},\{s_{i,t+k}\}_{k=1}^{m}$ a corresponding $Q$ update is based on the $Q$-value discrepancy given by:
	\begin{equation}
		\Delta Q_{i} = \frac{\|Q_{i}(\hat{s}_{i,t+k};\hat{\theta}) - Q_{i}(s_{i,t+k};\hat{\theta}) \|_{2}}{|\max\limits_{a,j}Q(\hat{s}_{i,t+j};\hat{\theta})_{a}|}
	\end{equation}
	where the $\hat{s}$ represents internal estimate of hidden state and $\hat{\theta}$ are the parameterized weights for the neural network approximating $Q_{i}$. As motivated in \cite{Kapturowski18} this helps measure the impact of representational drift and recurrent state staleness. The temporal differencing (TD) error in approximations of derived policy value represents the stepped difference between improvements toward true optimal value $V^*$ in \eqref{eqn:agentPol}. This is used internally by R2D2 to weight the state-action-reward sequences in the experience replay buffer. We set our corresponding priority exponent as $p=.6$, lower than in \cite{Kapturowski18}. This was due to the $n=1$ step update used for $Q$ reducing the averaging effect over long sequences of observations seen in \cite{Kapturowski18}. The choice of $n=1$ was motivated by considerations for training on tumbling modes where state estimation becomes exceedingly difficult on long time horizons. This similarly influenced our choice of agent reward discounting where we set $\gamma=.95$, 4.5 basis points lower than default. Finalized hyperparameter values are summarized in Table~\ref{table:Problem hyperparameters} below.
	
	\begin{table}[ht]
		\centering
		\begin{tabular}{|c|c|}
			\hline
			High-level Environment Parameters & Values\\
			\hline
			$\alpha$ & 2\\
			$\beta$ & 1\\
			$r_{0}$ & 0\\
			$\gamma$ & .95\\
			$M$ & .85\\
			\hline
			Hierarchical Environment Parameters & Values\\
			\hline
			$\epsilon$ & .35 m\\
			$\overline{t}$ & 50 s\\
			$\Delta t$ & 1s\\
			\hline
			R2D2 Hyperparameters &  \\
			\hline
			Learning Rate & 5e-5 \\
			Batch Size & 256\\
			Replay Burn-In & 20\\
			Replay Buffer Capacity & 100000\\
			Policy Max Seq Length (excl. burn-in) & 20\\
			Priority Exponent & .6\\
			Minibatch size & 128 \\
			Trained Iterations & 100000 \\
			\hline
			Network Parameters & \\
			\hline
			Type & Fully Connected, 2 layer, [64,64] \\
			LSTM Cell Size & 64 \\
			\hline
		\end{tabular}
		\caption{Hyperparameters and policy network for the multi-agent inspection problem.}
		\label{table:Problem hyperparameters}
	\end{table}
	
	The choice of reward shape was initially motivated by a simple metric for a linear combination between information gain and fuel cost of the form $r(s,a)=\alpha\frac{\|\mathbf{p}(t_{i,k+1})\land\neg \mathbf{p}(t_{i,k})\|_{1}}{|\cP|} + \beta \Delta V(\mathbf{x}_{k}^{i},\mathbf{x}_{k+1}^{i}) + r_{0}$. Unfortunately, the weight of intertemporal choice disincentivized agent movement when the feasible set of expectable newly visible POIs shrinks during the inspection mission. During early inspection stages, agents were eager to search for new information; at later inspection stages (when the set of unobserved POIs shrinks), each agent placed increased value on fuel optimal choices disregarding the remaining information retrieval criterion. To remedy this, we introduce the form in eq. \eqref{eqn:reward} offering a \emph{temporally} consistent incentive to look for the remaining POIs. For differing tuning parameters $\alpha,\beta,r_{0}$, the histogram shape during each episode reflects actions that heavily bias exploration vs exploitation. To address this, we set $\alpha,\beta,r_{0}$ to reflect an approximate median at $r(s,a)=0$; these are reported in Table~\ref{table:Problem hyperparameters}.

	\subsection{Navigation Control and Distributed Training}
	
	The full hierarchical inspection problem, consisting of both high level planners and low level point to point navigation controllers would be computationally burdensome to train on, due to the mixed temporal scale of the high level and low level actions and the necessity of full forward simulations of the physical agent trajectories. Though instances of long-horizon multi-agent games have been successfully trained, these often have simple action spaces and consistent temporal scales between actions and environment evolution, for instance \cite{Jaderberg19}. Instead, we looked to train the high level planner independently on its own abbreviated training environment (Problem \ref{prob:VSP}), then transfer the learned policies to the hierarchical environment to be deployed jointly with the low level navigation controllers. The hierarchical inspection environment thus consist of three high-low level agent pairs. Steps were taken to ensure that the high level and hierarchical environments have similar decision spaces for the view point planner to act on. Broadly, the MPC controller for point-to-point navigation was tuned to ensure consistency with the implied trajectories in the high level environment, and care was taken to train independent distributed view-point planning policies so that they exhibit similar behaviours when deployed on the asynchronous independent agents in the hierarchical inspection environment.
	
	We needed to ensure that information leakage in the reward function wasn't a problem during hierarchical rollout. This was what informed the imposition of two strong assumptions on the low-level controller. We required that TOF \eqref{eqn:LL_proxy} and fuel cost be statistically similar to the high-level trained versions, with an aggregate error of $<$10\% of high-level cost. This was enforced utilizing a constraint on target velocity where agreement between controllers (see Problem~\ref{prob: OptimizatonLLC}) is guaranteed through positional and temporal agreement upon switching controllers through the parameters $\epsilon$ and $\overline{t}$.  Since we aren't using RL methods for policy estimation of the low-level (as in \cite{Lei22}), we can partially ensure the low-level agents behave in a way that is correct-by-construction. Furthermore, this framework reflects consistency with the work established in \cite{Morgan14,Foust20} and can be directly analyzed for optimality within existing frameworks of guidance, navigation, and control. An MPC formulation of Problem~\ref{prob: OptimizatonLLC} is solved using sequential least squares quadratic programming (SLSQP) within Scipy's optimization framework. In this, we take a temporally greedy search strategy and iteratively optimize the $j^{\text{th}}$ step of Problem~\ref{prob: OptimizatonLLC} for each $j\in\{0,1,\dots,n-1\}$. The parameters $\epsilon$ and $\overline{t}$ are jointly set to ensure consistency (up to 10\% error) with the proxy values used during training of the high-level planner. Note that a single-burn controller is insufficient for this task due to accumulated spatial drifting as a result of the one step burn lag. Tuned parameter values are summarized in Table~\ref{table:Problem hyperparameters} above.
	
	Another key modeling decision is in the explicit characterization of the relationship between joint policies and agent-specific policies. Partially determined by joint reward structuring, the RL algorithm may be trained on a distributed single policy where the joint Q-function is learned and each agent's derived $Q_{i}$ function is homogeneously determined. In this scenario, joint policies $\pi = (\pi_{1},\pi_{2},\dots,\pi_{m})$ take the form $\pi = (\pi^*,\pi^*,\dots,\pi^*)$. This suggests that all agents make decisions according to the same policy mapping, albeit under fundamentally different sequences of observations. Heuristically, this represents training directly on the joint policy $\pi$ versus agent-specific policies used to \emph{construct} $\pi$. A clear advantage to this approach is in the model simplicity and homogeneity in policy behavior. In transfer learning, such policies may be less sensitive to perturbations in the underlying environment \cite{Torrey10}; such as training on a stable tumble and rolling out on a chaotic tumble. However, we found it necessary to train \emph{independent} policies due to the timing structure of our problem and the choice to treat the high-level independently of the low-level. Without additional information provided by the low-level (specifically in the time-step scale) the trained policies don't contain enough variance within sampling distribution to effectively explore the environment; even in the least temporally complicated case where the target is Nadir pointing. Training independent policies helps delineate choices through agent specific context and yields distributions that are statistically similar, but map to differing sets of actions; see row 3 of Table~\ref{tab:table1} in Section~\ref{subsec:Hier}. 

	\FloatBarrier
	\section{Results}
	\label{section:Results}
	We trained five different policies for each of the sample dynamics modes presented in Figure~\ref{fig:CombinedDynamics}. These all share the environmental and hyperparameter configuration presented in Table~\ref{table:Problem hyperparameters}. The resulting multi-agent policies were capable of inspecting a moving target across all 5 dynamic modes. They exhibited behaviours that intuitively represent notions of cooperation and fuel minimization while completing their target coverage objective. The high level policies were transferred to a hierarchical environment and similarly demonstrated success in completing the inspection problem cooperatively and with minimal fuel expenditure. A summary of the training process is shown in Figures \ref{fig:tboard_q} and \ref{fig:tboard_r}, and example trajectories and summary statistics for the hierarchical inspection problem are shown in Figures \ref{fig:insp_traj},  \ref{fig:rollout_traj} and Table~\ref{tab:table1}. Our results are presented in two sections; high level training and hierarchical policy rollout. Section~\ref{subsec:Training} presents results of the high-level training to solve Problem~\ref{prob:VSP} while Section~\ref{subsec:Hier} presents summary statistics and comparative performance for the full hierarchical rollout on each of the trained dynamic modes.
	
	\subsection{High Level Training Results}\label{subsec:Training}
	
	
	
	The high-level environment for different target rotation modes has to contend with two competing sources of uncertainty: the scale of $Q_{i}$ maximizing action spaces and the relative difficulty of hidden state estimation. There is less contributable variance between hidden state changes for static or slowly tumbling targets. This allows a larger sampling of observations around clusters of hidden state behavior providing more robust implicit state estimation. In other words, we expect the trained high-level policies to more easily contextualize the environment through observation. With that being said, these dynamic modes also have very particular viewpoint configuration graphs that can lead to maximal information discrimination. In this, we expect larger initial variability in estimates for state-action policy $Q_{i}$ that reduces much more quickly than the harder to characterize modes such as Stable Tumble and Chaotic Tumble. The assumptions present in Problem~\ref{prob:VSP} together with the configuration in Table~\ref{table:Problem hyperparameters} led to a sufficiently diverse space of rewards over observation, action pairs for R2D2 to effectively distinguish maximizers of each $Q_{i}$-function. The joint $Q$ function is then approximated through a particular learned tuple $(Q_{1}^{*},Q_{2}^{*},Q_{3}^{*})$. Note that this being additive over agent rewards suggests the trained policies represent single points on a hyperplane of maximizing $Q_{i}$ functions. As such, the space of solutions to $\pi_{i}^{*} \in \argmax_{\pi_{i}}V_{i}^{\pi_{i}}(\overline{o}_{i})$ is not unique. The plot in Figure~\ref{fig:tboard_q} below shows extreme values for $Q_{i}$ during training of two distinctly different rotation modes, Static Hill and a Single-axis Tumble; for more detail see Section~\ref{subsec:Hier}.

	\begin{figure}[h!]
		\centering
		\includegraphics[width=.75\linewidth]{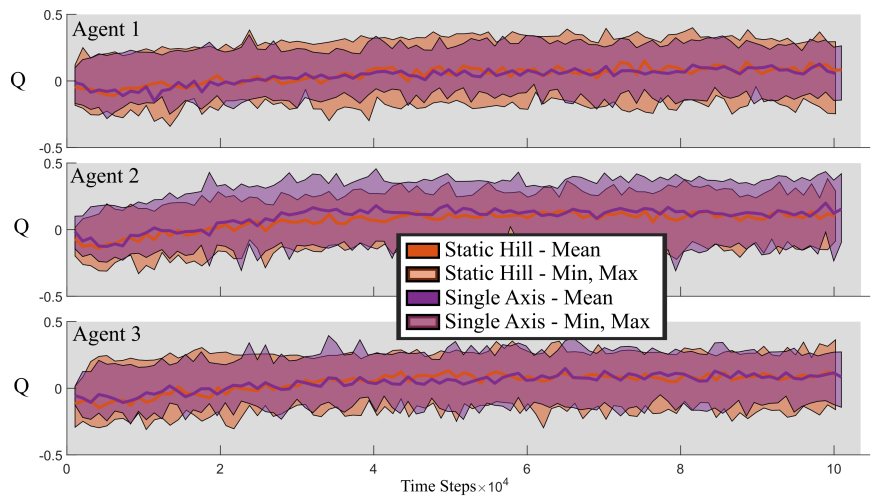}
		\caption{Q function estimation.}
		\label{fig:tboard_q}
	\end{figure}
	
\begin{figure}[h!]
		\centering
		\includegraphics[width=.65\linewidth]{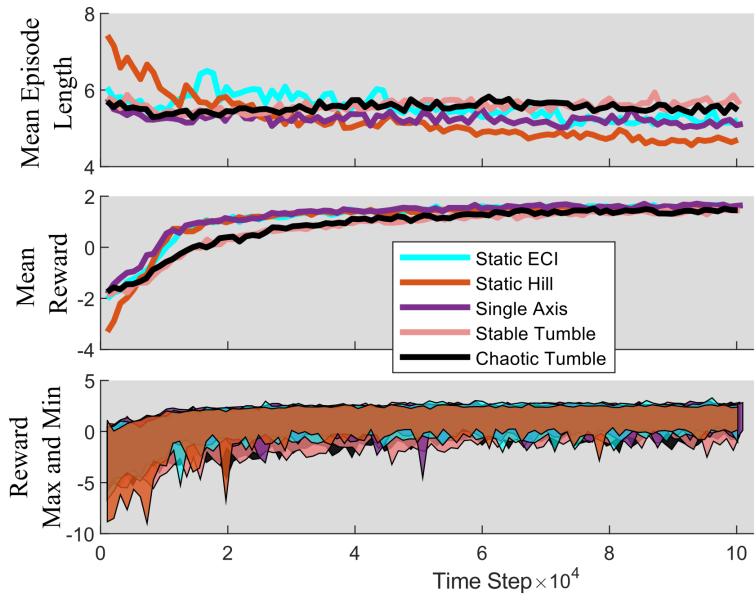}
		\caption{Policy reward and length distributions during training.}
		\label{fig:tboard_r}
	\end{figure}

	It is interesting to observe the induced variance as a result of poor state estimation through imperfect observations and uncertainty in hidden state estimation. The variance about the extended moving average (EMA) for each of the dynamic modes in Figure~\ref{fig:tboard_q} indicates a range of feasible actions that may result in comparably large rewards. In particular, for those modes with very well defined maximizing actions in conjunction with easier state estimation, the corresponding measure of variance about the estimated Q function is lower; for instance with the single axis rotation. A larger variance is partially indicative of a large set of actions providing similar rewards; thereby creating state-action ambiguity within training. In these cases, the learning rate is correspondingly slower; for instance in the tumbling modes in Figure~\ref{fig:tboard_r}\footnote{This is more directly evidenced in temporal difference error and rate of change in mean $Q_{i}$ across modes. For the sake of clarity, we did not include this in Fig.~\ref{fig:tboard_q}, but all data is available by request.}. Additionally, the advantage of training three independent policies for each mode versus a single shared policy (as in \cite{Lei22}) is reflected in the heterogeneity in learned $Q_{i}$. Each agent in the Static Hill mode has a similar internal reconstruction of $Q_{i}$ indicating homogeneity in the choice to explore or exploit environmental information; Table~\ref{tab:table1}. In contrast, taking advantage of learned geometry for the single axis rotation, $Q_{2}$ under agent-2 has dominating peaks and a higher average than its peers. This indicates a preference to explore new views in search of information gain whereas the other two indicate a willingness to exploit fuel-efficient view options; see Figure~\ref{fig:tboard_q}. 

	These metrics are directly related to observed mean episode length and episode reward (quoted for the joint policy); displayed in Figure~\ref{fig:tboard_r}. The mean episode length compiles an average number of high-level environment steps taken to reach inspection completion where the mean reward is calculated on instantaneous joint reward per episode step. These provide additional insight regarding the difficulty of learning movement behavior. In particular, both multi-axis tumbling modes exhibit a larger relative reward spread when compared to static and single-axis modes. The rate of change in reward mean is similarly slowed with training on both tumbles taking the longest to hit a target of 1.5. The mean episode length is partially correlated with parking actions. The Static Hill mode requires the fewest number of steps primarily due to forced exploration for information retrieval purposes. Parking, although relatively cheap will not result in any new information. All dynamic modes learn that parking may be a beneficial action and split behavior to hedge risk in the uncertainty induced through hidden state estimation; thereby increasing the required number of environment steps.  This indicates a larger number of viewpoint transfers leveraging target rotation moving POIs into the agent's FOV over time.

	\subsection{Hierarchical Policy Rollout}\label{subsec:Hier}
	
\begin{figure}[h!]
		\centering
		\includegraphics[width=1\linewidth]{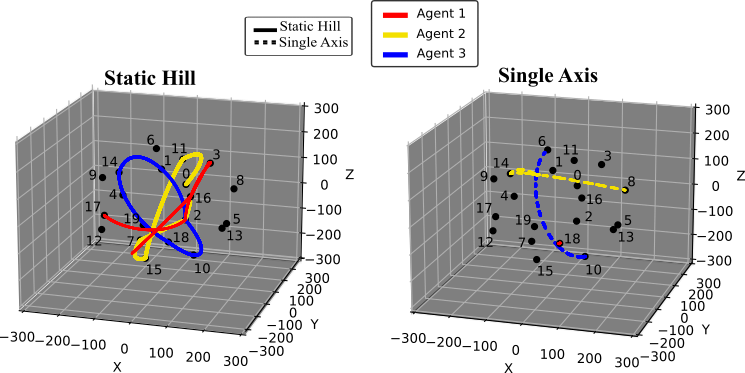}
		\caption{Agent trajectories for a hierarchical inspection mission under the Static Hill and Single-axis Tumble target rotation modes.}
		\label{fig:insp_traj}
	\end{figure}

	To test the hierarchical approach, we simulated the full environment as described in Section~\ref{subsection:mpc_prob}, using the trained high-level policies as trained and tested above in combination with the MPC low-level controller. Although further policy evaluation is required to test for \textit{statistically significant} differences within agent-level inter-mode policies, a sample of $100$ inspection runs for each rotation mode (see Table~\ref{tab:table1}) suggests key differences in \emph{joint} strategy development. Our relatively simple high-level environment increases the likelihood of causal explanation originating from the change in underlying target dynamic mode. High-level trajectory analysis and stability of estimates for agent specific policy mean and variance indicate diversity in learned strategy within each mode across the set of \emph{agents}. As an example, we have provided a hierarchical multi-agent inspection trajectory/rollout for two distinct dynamic modes in Figure~\ref{fig:insp_traj}.

	\begin{figure}[h!]
		\centering
		\includegraphics[width=1\linewidth]{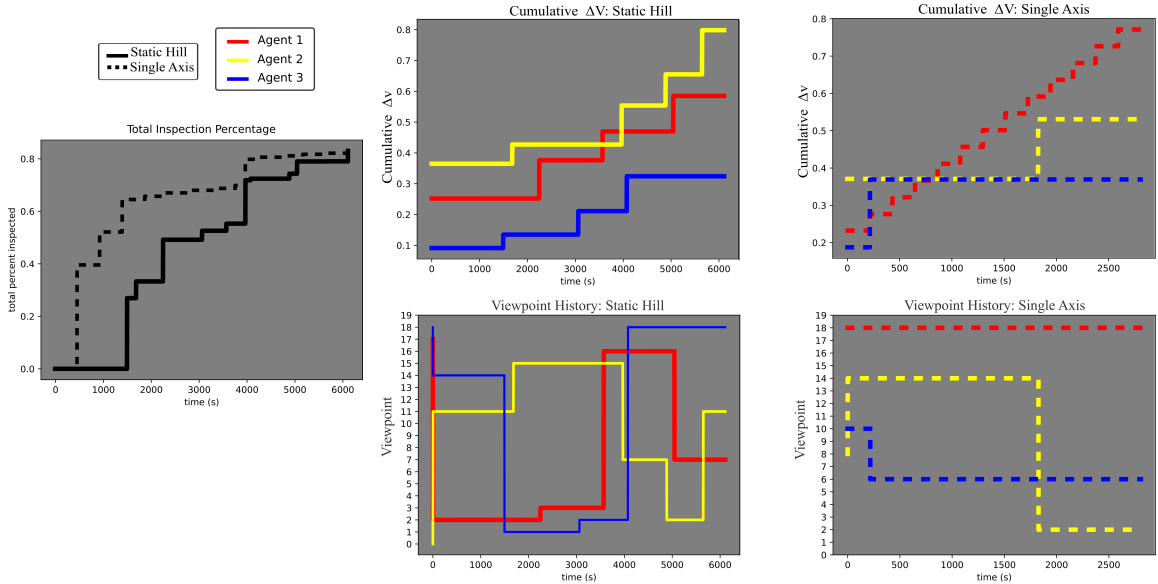}
		\caption{Inspection percentage, cumulative fuel cost, and viewpoint transfers plotted against simulation time for the Static Hill and Single Axis rotation modes.}
		\label{fig:rollout_traj}
	\end{figure}
	
	\begin{table}[h!]
		\begin{center}
			\begin{tabular}{l|c|c|c|c|c} 
				\textbf{Mean Metrics}  & \textbf{Static Hill} & \textbf{Static ECI} & \textbf{Single Axis} & \textbf{Stable Tumble} & \textbf{Chaotic Tumble}\\
				\hline
				Inspection \%  & 83.28 & 83.06 & 83.63 & 83.03 & 83.03\\
				Time (s)  & 6625.06 & 4751.51 & 3641.69 & 5228.71 & 3910.14\\
				Actions $\frac{Unique}{Total}$ 
				& $\frac{3.79}{4.67}$,$\frac{4.08}{4.83}$,$\frac{3.96}{4.68}$
				& $\frac{1.53}{13.45}$,$\frac{2.52}{4.47}$,$\frac{2.11}{8.35}$ 
				& $\frac{1.16}{10.46}$,$\frac{2.42}{2.87}$,$\frac{2.34}{2.98}$
				& $\frac{2.57}{6.08}$,$\frac{2.87}{6.10}$,$\frac{3.34}{5.09}$
				& $\frac{2.05}{7.05}$,$\frac{2.56}{3.89}$,$\frac{2.40}{5.27}$\\
				Accum. $\Delta$V  (m/s) & 2.08 & 2.35 & 1.74 & 3.20 & 2.55\\
			\end{tabular}
		\end{center}
		\caption{Mean hierarchical rollout metrics by dynamic mode over 100 runs. The actions metric is given as a tuple indicating the average proportion of unique to total actions taken during an inspection episode for each of the three agents. The policies were trained to inspect 85\% of $\cP$; due to the asynchronous agent movement, we rolled out on an 83\% threshold for this table.}
		\label{tab:table1}
	\end{table}
	
	\subsubsection{Static Hill: }
	For this case, we fix target rotation mode by taking initial target angular velocity as $\omega^{BF} = [0, 0, n]$ where $n = 0.001027$ rad/s is the target's mean motion corresponding to a static target in Hill's frame. This implies that parking actions will yield no new information from the target. As captured in Figure~\ref{fig:insp_traj}, agents inspecting this target chose to navigate to points along a trajectory providing maximally discriminative viewpoints of the target. Note that the geometric diversity is partially obfuscated by the nonconvex surface of the target. In particular, viewpoints that are relatively close in Hill's frame are still geometrically optimal due to POIs hidden by satellite antennae and communications arrays; see Figure~\ref{fig:AuraInsEx} for further intuition. Agent specific policy distributions are shifted forms of one another with differences in probability mass concentrated around specific regions of the viewpoint graph. The separation here further exemplifies the necessity of training policies for each agent versus reliance on a single, distributed policy as in \cite{Lei22}. The inspection took the longest of all the modes primarily due to the transfer time imposed by \eqref{eqn:LL_proxy} and the suboptimality of vehicle parking. This is reflected in the homogeneity between agents when comparing the ratio between unique and total actions required for inspection, see the third row of Table~\ref{tab:table1}.  A distinctive feature for this rotation mode is the lower number of average environment steps needed to complete inspection during rollout. This indicates that agents are effectively learning to strategically explore the viewpoint graph based on single transfers versus a string of multiple transfers. The reduced variance in agent-policy mappings help ensure that specific inspection trajectories have a higher likelihood of being fuel efficient relative to other dynamic modes; see for instance cumulative $\Delta V$ in Figure~\ref{fig:rollout_traj}.
	
	\subsubsection{Static ECI and Single-Axis: }
	Both of these cases represent a single-axis rotation about the z-axis. Here $\omega^{BF} = [0, 0, 0]$ and $\omega^{BF} = [0, 0, .097]$ for which the scale factor $.097$ is indicative of the number of $2\pi$ rotations performed about the z-axis for each target orbital period; given by $.097/n\approx 94.45$ rotations. These two modes represent rotations with opposing direction and differing rotation rate.  The exhibited dynamic behavior is interesting because it induces nontrivial state estimation and provides a strong intuition originating from geometrical arguments. Broadly, we expected agents to attempt movements in a plane whose normal vector \emph{nearly} coincides with the axis of rotation. It should not be parallel owing to the target convexity and initial orientation. Angular direction should be opposite of target rotation direction, heuristically letting points fall into view on their own accord.
	
	The resulting trained strategies are similar in nature. Depending on the initial locations of the agents, agent 1 learned to perpetually stay at the same point providing a reasonable balance between information gain and fuel efficiency. More importantly, increasing the predictability of the inspection time horizon (remember that agents can't explicitly share inspection progress) allowing for agent 1 and agent 2 to effectively exploit fuel efficient, information discriminative views. Agent 2 moves for immediate information gain whereas agent 3 moves for immediate fuel conservation. This plays out at two different time scales for each rotation mode. For the single-axis case, one agent favors parking; the other two agents only have one or two feasible actions before the inspection finishes due to rotation rate. In this case, Agent 2 moves to cover the target caps (which Agent 1 can't witness while parked) and Agent 3 seeks a fuel efficient viewpoint to park on, see the far right panel in Figure~\ref{fig:rollout_traj}. The parking nature of Agent 3 is exemplified in the third row of Table~\ref{tab:table1} where you can see a much lower ratio between unique to total viewpoint actions taken when compared to single-axis. This is largely attributable to rotation rate in which less information rotates into view between parking actions, extending time horizon. Referencing the fuel costs in Figure~\ref{fig:rollout_traj}, note that Agent 1 accumulates $\Delta V$ according to a step function consistent with the calculation made in \eqref{eqn:delV} and Problem~\ref{prob: OptimizatonLLC}. It is important to note that the zero-fuel cost incurred by Agent 3 reflects accrual during the transition between viewpoint 10 and viewpoint 6; not a parking action. These modes broadly demonstrate the capability of \textit{indirect} multi-agent coordination learned through action contextualization of peers in the absence of action telegraphing or \emph{direct} coordination.

	\subsubsection{Stable and Chaotic Tumble: }
	These two modes represent multi-axis tumbles. Here, we fix the rotation mode by taking initial target angular velocity as $\omega^{BF} = [0.0097, .097, 0]$ for a stable tumble (y-axis with x-axis perturbation) or  $\omega^{BF} = [0.0097, 0, .097]$ for an unstable tumble (z-axis with x-axis perturbation). The stable and then chaotic tumble represents an increasingly large solution space of feasible optimal viewpoint candidates. This makes both exploration and exploitation equally enticing to each agent; similar to Static Hill. Unlike Static Hill, parking actions are now valuable due to the decreased likelihood of witnessing the same POIs from two different points in time and space. This makes hidden state estimation increasingly difficult and creates strategies that have a similar propensity to maintain geometrically consistent spatial inspection patterns. The key differentiation between single-axis rotations and these two modes is in the complexity of dynamic motion coupled with the number of efficacious actions. With an increased propensity to park, see the third row of Table~\ref{tab:table1}, the inspection time horizon is shorter than in Static Hill. Unintuitively, this doesn't decrease the average cumulative $\Delta V$ required for inspection. This is because there is increased unpredictability in tail-end retrieval due to the difficult hidden state estimation. As the inspection task nears its end, there is a large spike in fuel burn expended to intercept the remaining unseen POIs. Unlike Static Hill, advantageous \textit{future} positions on long time-scales aren't easily predictable increasing the tendency of agents to weight fuel lower than information gain in policy distribution. Moreover, the impact of surface geometry on inspection difficulty can also be seen in the differential between accumulated $\Delta V$ and inspection time in Table~\ref{tab:table1}. On average, the chaotic tumble took only 80\% of the required fuel and 75\% of the required time compared with the stable tumble. This originates from the primary axis of rotation used in the stable tumble mode. More specifically, rotation about the y-axis relegates priority viewpoint transfers to off-axis positions thereby forcing fuel expensive decisions.

	\section{Discussion and Concluding Remarks}
	\label{section:Conclusions}
	
	In this paper, we considered a hierarchical approach for the decentralized planning of multi-agent inspection of a tumbling target. This consisted of two components; a viewpoint or high-level planner trained using deep reinforcement learning and a navigation planner handling point-to-point navigation between pre-specified viewpoints. Operating under limited information, our trained multi-agent high-level policies successfully contextualize information within the global hierarchical environment and are correspondingly able to inspect over 90\% of non-convex tumbling targets (corresponding to 85\% of the actual POIs in $\cP$), even in the absence of additional agent attitude control. Our assumption that viewpoint traversal coincides with camera shutter led to artificially long inspection missions; this can be ameliorated by increasing viewpoint density and sampling from approximate neighborhoods utilizing action masking. This will substantially increase offline training time, but will have negligible effect on policy rollout and online decision making. 
	
	Our results show promise in the incorporation of machine-learning based methods for autonomous inspection tasks. For viewpoint planning in the face of partial information and limited observability, our agents can quickly and efficiently contextualize environmental state with action mapping in numerous dynamic scenarios. We tested this under five modes: Static Hill (Nadir pointing), Static ECI (star pointing), single-axis rotation, stable tumble, and chaotic tumble. Each of these helped demonstrate the diversity in inspection strategy witnessed as a result of the direct modeling of information-gain based agent reward. We sampled all hierarchical policies in each mode to get estimates of key inspection statistics; these revealed large differences in learned strategy. Agents inspecting a (relatively) static target learn to move to distinct viewpoints that maximize information discrimination with minimal fuel cost. Contrasting this is the case of single-axis rotation and Static ECI where agents learn to park and take advantage of rotational dynamics to observe new information with minimal fuel penalty. The variability in rolled-out policies for chaotic and stable tumbles together with higher cumulative $\Delta V$ as compared with simpler modes indicate difficulty in finding unseen POIs near the end of mission. This effect may be reduced through reward and hyperparameter tuning or through stronger assumptions on agent observations. The variety in learned behavior indicates that the potential of transfer learning between modes may be limited; although a wider sampling of dynamic modes may be used to extend the observation and state space within the DEC-POMDP formulation. In this context, competing solutions (such as \cite{Lei22}) that make assumptions on the validity of information retrieval do not provide any real guarantee of inspectability, even through complete graph traversal. Our formulation partially demonstrates that such paradigms may increase fuel cost unnecessarily by requiring visitation at all viewpoints regardless of inspection state. In comparison with dedicated computer-vision based techniques such as ORB-SLAM, our framework directly incorporates \textit{path planning and optimization} on the basis of implicit belief state versus assuming a fixed trajectory for path planning \text{and} navigation.
	
	Future work aims to investigate the potential for model scalability through increasingly complicated modeling assumptions and technicalities. For instance potentially supporting uncertainty in target geometry and rotation dynamics estimation, torque induced dynamical modes, safety constraints, attitude control, elliptical NMT viewpoint discretization, communications dropouts, variable camera shutter rate, and hybrid mutual information metrics. Furthermore, it would be prudent to extend analysis to directly compare the effect of training a high-level policy directly versus the full hierarchical environment. Depending on model complexity and the density of sample observations, it may be necessary to train the hierarchical agents directly. This may help refine internal belief state estimation in the face of increasing variance as a result of observation ambiguity. Lastly, a more robust and empirically motivated low-level navigation controller can be integrated during training.

	\bibliographystyle{ieeetr}
	\bibliography{references}
	
\end{document}